\documentclass[11pt]{article}

\usepackage{naacl2021}

\usepackage{times}
\usepackage{latexsym}

\usepackage[T1]{fontenc}
\usepackage[utf8]{inputenc}

\usepackage{microtype}

\usepackage{url}
\usepackage{color}
\usepackage{multirow}
\usepackage{subfig}
\usepackage{arydshln}
\usepackage{CJKutf8}

\newcommand{\red}[1]{\textcolor[gray]{0.4}{#1}}
\newcommand{\ja}[1]{\begin{CJK}{UTF8}{ipxm}#1\end{CJK}}
\newcommand{\zh}[1]{\begin{CJK}{UTF8}{gbsn}#1\end{CJK}}

\title{User-Generated Text Corpus for Evaluating Japanese Morphological Analysis and Lexical Normalization}

\author{
Shohei Higashiyama${}^{1,2}$, Masao Utiyama${}^1$, Taro Watanabe${}^2$, Eiichiro Sumita${}^1$\\
  ${}^1$National Institute of Information and Communications Technology, Kyoto, Japan \\
  ${}^2$Nara Institute of Science and Technology, Nara, Japan \\
  {\tt \{shohei.higashiyama, mutiyama, eiichiro.sumita\}@nict.go.jp,}\\
  {\tt taro@is.naist.jp}
}

\begin{document}
\maketitle
\begin{abstract}
Morphological analysis (MA) and lexical normalization (LN) are 
both important tasks for Japanese user-generated text (UGT).
To evaluate and compare different MA/LN systems,
we have constructed a publicly available Japanese UGT corpus.
Our corpus comprises 929 sentences annotated with 
morphological and normalization information,
along with category information we classified for frequent UGT-specific phenomena.
Experiments on the corpus demonstrated 
the low performance of existing MA/LN methods for non-general words and non-standard forms,
indicating that the corpus would be a challenging benchmark for further research on UGT.
% can be a major challenge in UGT.
\end{abstract}

\section{Introduction}
Japanese morphological analysis (MA) is a fundamental and important task 
that involves word segmentation, part-of-speech (POS) tagging and lemmatization
because the Japanese language has no explicit word delimiters.
Although MA methods for well-formed text \cite{kudo2004,neubig2011}
have been actively developed taking advantage of the existing annotated corpora of 
newswire domains,
they perform poorly on user-generated text (UGT), such as social media posts and blogs.
Additionally, because of the frequent occurrence of informal words, 
lexical normalization (LN), which identifies standard word forms, is another important task in UGT.
Several studies have been devoted to both tasks in Japanese UGT
\cite{sasano2013,kaji2014,saito2014a,saito2017a}
to achieve the robust performance for noisy text.
Previous researchers have evaluated their own systems using in-house data created by individual researchers,
and thus it is difficult to compare the performance of different systems and
discuss what issues remain in these two tasks.
Therefore, publicly available data is necessary for a fair evaluation of
MA and LN performance on Japanese UGT.

In this paper, we present the blog and Q\&A site normalization corpus (BQNC),\footnote{Our corpus will be available at \url{https://github.com/shigashiyama/jlexnorm}~.}
which is a public Japanese UGT corpus annotated with morphological and normalization information.
%word boundaries, POS, lemma, and standard forms of non-standard words.
We have constructed the corpus under the following policies:
(1) available and restorable;
(2) compatible with the segmentation standard and POS tags used in 
the existing representative corpora; and
(3) enabling a detailed evaluation of UGT-specific problems.

For the first requirement,
we extracted and used the raw sentences in the blog and Q\&A site registers compiled by
(the non-core data of) the Balanced Corpus of Contemporary Written Japanese (BCCWJ) \cite{maekawa2014},
in which the original sentences are preserved.\footnote{Twitter could be a candidate for a data source.
However, redistributing original tweets collected via the Twitter Streaming APIs
is not permitted by Twitter, Inc., and an alternative approach to distributing tweet URLs
has the disadvantage that the original tweets can be removed in the future.}
% Anyone can restore the complete data of the BQNC by combining
% the original BCCWJ data and 
For the second requirement, we followed the short unit word (SUW) criterion of
the National Institute for Japanese Language and Linguistics (NINJAL), 
which is used in various NINJAL's corpora,
including manually annotated sentences in the BCCWJ.
For the third requirement, we organized linguistic phenomena frequently observed in the two registers
as word categories, and annotated each word with a category.
We expect that this will contribute to future research 
to develop systems that manage UGT-specific problems.

\begin{table*}[t]
 \centering
 \footnotesize
 \begin{tabular}{llllll} \hline \hline
  & Category                    & Example  & Reading & Translation & Standard forms \\
  \hline
  \multirow{8}{*}{Type of vocabulary}
  & (General words)             & \\
  & Neologisms/Slang            & \ja{コピペ}   & \textit{copipe}  & copy and paste & \\
  & Proper names                & \ja{ドラクエ} & \textit{dorakue} & Dragon Quest & \\
  & Onomatopoeia                & \ja{キラキラ} & \textit{kirakira}& glitter     & \\
  & Interjections               & \ja{おお}     & \textit{\=o}     & oops        & \\ 
  & Dialect words               & \ja{ほんま}   & \textit{homma}   & truly       & \\
  & Foreign words               & \ja{ＥＡＳＹ} && easy                          & \\
  & Emoticons/AA                & \ja{（＾−＾）} &&& \\
  \hline
  \multirow{5}{*}{Type of variant form}
  & (Standard forms)            & \\
  & Character type variants     & \ja{カワイイ} & \textit{kawa\={\i}} & cute  & \ja{かわいい,可愛い}   \\
  & Alternative representations & \ja{大きぃ}     & \textit{\=ok\=i}   & big   & \ja{大きい}            \\
  & Sound change variants       & \ja{おいしーい} & \textit{oish\={\i}i}& tasty & \ja{おいしい,美味しい} \\
  & Typographical errors        & \ja{つたい}     & \textit{tsutai}  & tough & \ja{つらい,辛い}       \\
  \hline
 \end{tabular}
 \caption{Word categories in the BQNC} \label{tab:cate}
\end{table*}

The BQNC comprises sentence IDs and annotation information, including
word boundaries, POS, lemmas, standard forms of non-standard word tokens, and word categories.
We will release the annotation information that enables BCCWJ applicants
to replicate the full BQNC data from the original BCCWJ data.%\footnote{\url{https://pj.ninjal.ac.jp/corpus_center/bccwj/en/subscription.html}}

Using the BQNC, we evaluated two existing methods:
a popular Japanese MA toolkit called MeCab \cite{kudo2004} 
and a joint MA and LN method \cite{sasano2013}.
Our experiments and error analysis showed that 
these systems did not achieve satisfactory performance for non-general words.
This indicates that our corpus would be a challenging benchmark for further research on UGT.

\section{Overview of Word Categories} \label{sec:cate}
Based on our observations and the existing studies \cite{ikeda2010,kaji2015},
we organized word tokens that may often cause segmentation errors
into two major types with several categories, as shown in Table \ref{tab:cate}.
We classified each word token from two perspectives:
the type of vocabulary to which it belongs and the type of variant form to which it corresponds.
For example, \ja{ニホン} \textit{nihon} `Japan' written in \textit{katakana}
corresponds to a \textit{proper name} and a \textit{character type variant} 
of its standard form \ja{日本} written in \textit{kanji}.

Specifically, we classified vocabulary types into 
\textit{neologisms/slang}, \textit{proper names}, \textit{onomatopoeia},\footnote{``Onomatopoeia'' 
typically refers to both the phonomime and phenomime
in Japanese linguistics literature, similar to ideophones. We follow this convention in this paper.} 
\textit{interjections}, \textit{(Japanese) dialect words}, \textit{foreign words}, and 
\textit{emoticons/ASCII art (AA)}, in addition to general words.\footnote{We observed 
a few examples of other vocabulary types,
such as Japanese archaic words and special sentence-final particles in our corpus,
but we treated them as general words.}
A common characteristic of these vocabularies, except for general words,
is that a new word can be indefinitely invented or imported.
We annotated word tokens with vocabulary type information, except for general words.

From another perspective, any word can have multiple variant forms.
Because the Japanese writing system comprises multiple script types
including \textit{kanji} and two types of \textit{kana}, that is,
\textit{hiragana} and \textit{katakana},\footnote{Morphographic \textit{kanji} and 
syllabographic \textit{hiragana} are primarily used for 
Japanese native words (\textit{wago}) and Japanese words of Chinese origin 
(Sino-Japanese words or \textit{kango}),
whereas syllabographic \textit{katakana} is primarily used, for example, 
for loanwords, onomatopoeia, and scientific names.
Additionally, Arabic numerals, Latin letters (\textit{r\=omaji}), 
and other auxiliary symbols are used in Japanese sentences.}
words have orthographic variants written in different scripts.
Among them, non-standard \textit{character type variants} that rarely occur in well-formed text but occur in UGT
can be problematic, for example, a non-standard form \ja{カワイイ} for a standard form \ja{かわいい} \textit{kawa\={\i}} `cute'.
Additionally, ill-spelled words are frequently produced in UGT.
We further divided them into two categories.
The first is \textit{sound change variants} that
have a phonetic difference from the original form
and are typically derived by deletions, insertions, or substitutions of
vowels, long sound symbols (\textit{ch\=oon} ``\ja{ー}''), long consonants (\textit{sokuon} ``\ja{っ}''), 
and moraic nasals (\textit{hatsuon} ``\ja{ん}''), for example, \ja{おいしーい} \textit{oish\={\i}i}
for \ja{おいしい} \textit{oish\={\i}} `tasty'.
The second category is \textit{alternative representations} that 
do not have a phonetic difference and are typically achieved by
substitution among uppercase or lowercase kana characters, or
among vowel characters and long sound symbols, for example, \ja{大きぃ} for \ja{大きい} \textit{\=ok\={\i}} `big'.
Moreover, \textit{typographical errors} can be seen as another type of variant form.
We targeted these four types of non-standard forms 
for normalization to standard forms.

\section{Corpus Construction Process}
The BQNC was constructed using the following steps.
The annotation process was performed by the first author.

\paragraph{(1) Sentence Selection}
We manually selected sentences to include in our corpus
from the blog and Q\&A site registers in the BCCWJ non-core data.
We preferentially extracted sentences that contained candidates of UGT-specific words,
that is, word tokens that may belong to non-general vocabularies or correspond to non-standard forms.
As a result, we collected more than 900 sentences.

\paragraph{(2) First Annotation}
Sentences in the non-core data have been automatically annotated
with word boundaries and word attributes, such as POS and lemma.
% \footnote{Automatic annotation was amed at achieving accuracy of 98\% \cite{ogura2011}.}
Following the BCCWJ annotation guidelines \cite{ogura2011b,ogura2011a} and UniDic \cite{den2007},
which is an electronic dictionary database designed for the construction of NINJAL's corpora,
we refined the original annotations of the selected sentences by manually checking them.
The refined attributes were token, POS, conjugation type, conjugation form, pronunciation, lemma, and lemma ID.
Additionally, we annotated each token with a word category shown in Table \ref{tab:cate} 
and a standard form ID if the token corresponded to a non-standard form.

Table \ref{tab:anno_exam} shows two examples of annotated sentences.
We annotated each non-standard token with a standard form ID denoted 
as ``[lemma ID]:[lemma](\_[pronunciation])'',
which is %defined for each word form and 
associated with the set of acceptable standard forms shown in Table \ref{tab:stdformid}.

\begin{table}[t]
 \centering
 \footnotesize
 \begin{tabular}{lll} \hline \hline
  Token& Translation            & Standard form ID \\ \hline
  \ja{イイ} & good      & 38988:\ja{良い} \\
  \ja{歌}   & song      & \\ 
  \ja{です} & (polite copula)   & \\ 
  \ja{ねェ} & (emphasis marker) & 28754:\ja{ね} \\
  \hline
  \ja{ヨカッ} & good    & 38988:\ja{良い\_ヨカッ}  \\ 
  \ja{タ}     & (past tense marker) & 21642:\ja{た} \\ 
  \hline
 \end{tabular}
 \caption{Examples of annotated text ``\ja{イイ歌ですねェ}'' (It's a good song, isn't it?)
 and ``\ja{ヨカッタ}'' (It was good.). 
 Attributes except for token and standard form ID are abbreviated.
 } \label{tab:anno_exam}
\end{table}

\begin{table}[t]
 \centering
 \footnotesize
 \begin{tabular}{ll} \hline \hline
  Standard form ID   & Standard forms \\ \hline
  21642:\ja{た}           & \ja{た} \\ 
  28754:\ja{ね}           & \ja{ね} \\ 
  38988:\ja{良い}         & \ja{良い,よい,いい} \\
  38988:\ja{良い\_ヨカッ} & \ja{良かっ,よかっ} \\ 
  \hline
 \end{tabular}
 \caption{Examples of standard form IDs} \label{tab:stdformid} 
\end{table}

\paragraph{(3) Second Annotation}
We rechecked all tokens in the sentences that we finished the first annotation
and fixed the annotation criteria, that is, the definitions of
vocabulary types and variant form types, and standard forms for each word.
Through these steps, we obtained 929 annotated sentences.

\section{Detailed Definition of Word Categories}
\subsection{Type of Vocabulary}
Through the annotation process, we defined the criteria for vocabulary types as follows.

\paragraph{Neologisms/Slang:}
a newly invented or imported word that has come to be used collectively.
Specifically, we used a corpus reference application called Chunagon\footnote{\url{https://chunagon.ninjal.ac.jp}}
and regarded a word as a \textit{neologism/slang}
if its frequency in the BCCWJ was less than five before the year 2000
and increased to more than ten in 2000 or later.\footnote{The original sentences were 
from posts published between 2004 and 2009.}

\paragraph{Proper names:}
following the BCCWJ guidelines,
we regarded a single word that corresponded to a proper name, such as 
person name, organization name, location name, and product name, as a \textit{proper name}.
In contrast to the BCCWJ guidelines,
we also regarded an abbreviation of a proper name as a \textit{proper name},
for example, ``\ja{ドラクエ}'' in Table \ref{tab:cate}.

\paragraph{Onomatopoeia:}
a word corresponds to onomatopoeia. 
We referred to a Japanese onomatopoeia dictionary \cite{yamaguchi2002} 
to assess whether a word is onomatopoeic.
We followed the criteria in the BCCWJ guidelines on
what forms of words are onomatopoeic and 
what words are associated with the same or different lemmas.

\paragraph{Interjections:}
a word whose POS corresponds to an interjection.
Although we defined standard forms for idiomatic greeting expressions
registered as single words in UniDic,\footnote{Eight greeting words exist, 
for example, \ja{ありがとう} \textit{arigat\=o} `thank you' and 
\ja{さようなら} \textit{say\=onara} `see you'.}
we did not define standard and non-standard forms for other interjections that
express feelings or reactions, 
for example, \ja{ええ} \textit{\=e} `uh-huh' and \ja{うわあ} \textit{uw\=a} `wow'.

\paragraph{Foreign words:}
a word from non-Japanese languages.
We regarded a word written in scripts in the original language as a \textit{foreign word},
for example, English words written in the Latin alphabet such as ``plastic''.
Conversely, we regarded loanwords written in Japanese scripts
%(\textit{hiragana}, \textit{katakana}, or \textit{kanji})
(hiragana, katakana, or kanji) as general words, for example, \ja{プラスチック} `plastic'.
Moreover, we did not regard English acronyms and 
abbreviations written in uppercase letters as foreign words
because such words are typically also written in the Latin alphabet in Japanese sentences, 
for example, \ja{ＳＮＳ}.
%``SNS''. % `SNS'.

\paragraph{Dialect words:} 
a word from a Japanese dialect.
We referred to a Japanese dialect dictionary \cite{sato2009}
and regarded a word as a \textit{dialect word} 
if it corresponded to an entry or occurred in an example sentence.
We did not consider normalization from a dialect word to 
a corresponding word in the standard Japanese dialect.

\paragraph{Emoticons/AA:}
nonverbal expressions that comprise characters to express feelings or attitudes.
Because the BCCWJ guidelines does not explicitly describe
criteria on how to segment emoticon/AA expressions as words,
we defined criteria to follow emoticon/AA entries in 
UniDic.\footnote{For example, if characters expressing body parts were
outside of punctuation expressing the outline of a face,
the face and body parts were segmented, but both were annotated with \textit{emoticons/AA},
for example, ``\ja{ｍ（．＿．）ｍ}'' $\rightarrow$ ``\ja{ｍ}$|$\ja{（．＿．）}$|$\ja{ｍ}''.}

\subsection{Type of Variant Form}
There are no trivial criteria to determine which variant forms of a word are standard forms
because most Japanese words can be written in multiple ways.
Therefore, we defined standard forms of a word
as all forms whose occurrence rates were approximately equal to 10\% or more in the BCCWJ
among forms that were associated with the same lemma.
For example, among variant forms of the lemma \ja{面白い} \textit{omoshiroi} `interesting' or `funny'
that occurred 7.9K times, 
major forms \ja{面白い} and \ja{おもしろい} accounted for 72\% and 27\%, respectively,
and other forms, such as \ja{オモシロイ} and \ja{オモシロい}, were very rare.
In this case, the standard forms of this word are the two former variants.
We annotated tokens corresponding to the two latter non-standard forms 
with the standard form IDs and the types of variant forms.
We defined criteria for types of variant forms as follows.

\paragraph{Character type variants:} 
among the variants written in different scripts,
we regarded variants whose occurrence rates were approximately equal to 5\% or less in the BCCWJ
as non-standard forms of \textit{character type variants}.
Specifically, 
variants written in kanji, hiragana, or katakana
for native words and Sino-Japanese words,
variants written in katakana or hiragana for loanwords,
variants written in uppercase or lowercase Latin letters for English abbreviations
are candidates for character type variants.
We assessed whether these candidates were non-standard forms based on the occurrence rates.

\paragraph{Alternative representations:}
a form whose internal characters are (partially) replaced by 
special characters without phonetic differences.
Specifically, non-standard forms of \textit{alternative representations} include
native words and Sino-Japanese words written in historical kana 
orthography (e.g., \ja{思ふ} for \ja{思う} \textit{om\=o}/\textit{omou} `think'),
and loanwords written as an unusual\footnote{We assessed whether a form is unusual
if its occurrence rate was approximately equal to 5\% or less in the BCCWJ
similar to the case of character type variants.} 
katakana sequence (e.g., \ja{オオケストラ} for \ja{オーケストラ} `orchestra').
Additionally, \textit{alternative representations} include substitution with respect to kana: 
substitution of the long vowel kana by the long sound symbol
(e.g., \ja{おいし〜} for \ja{おいしい} \textit{oish\={\i}} `tasty'),
substitution of upper/lowercase kana by the other case
(e.g., \ja{ゎたし} for \ja{わたし} \textit{watashi} `me'), and
phonetic or visual substitution of kana characters by Latin letters and symbols
(e.g., \ja{かわＥ} for \ja{かわいい} \textit{kawa\={\i}} `cute' and
\ja{こωにちは} for \ja{こんにちは} \textit{konnichiwa} `hello').

\paragraph{Sound change variants:}
a form whose pronunciation is changed from the original form.
Specifically, \textit{sound change variants} include
the insertion of special moras (e.g., \ja{強ーい} \textit{tsuy\=oi} for \ja{強い} \textit{tusyoi} `strong'),
deletion of moras (e.g., \ja{くさ} \textit{kusa} for \ja{くさい} \textit{kusai} `stinking'), and
substitution of characters/moras (e.g.,
\ja{っす} \textit{ssu} for \ja{です} \textit{desu} polite copula and
\ja{すげえ} \textit{sug\=e} for \ja{すごい} \textit{sugoi} `awesome').
%, and \ja{ぴったし} \textit{pittashi} for \ja{ぴったり} \textit{pittari} `exactly').

\paragraph{Typographical errors:}
a form with typographical errors derived from 
character input errors, kana-kanji conversion errors, or the user's incorrect understanding.
For example, \ja{つたい} \textit{tsutai} for \ja{つらい} \textit{turai} `tough'
and \ja{そｒ} for \ja{それ} \textit{sore} `it'.

\section{Evaluation} \label{sec:eval}
\begin{table}[t]
 \centering
 \footnotesize
 \begin{tabular}{l|rrrrr} \hline \hline
  Regi-& \# sent & \# word & \# word & \# NSW & \# NSW \\
  ster  &         & token   & type    & token  & type \\ \hline
  Q\&A  & 379 & 5,649 & 1,699 & 320 & 221 \\
  Blog  & 550 & 6,951 & 2,231 & 447 & 257 \\ \hline 
  Total & 929 & 12,600& 3,419 & 767 & 420 \\
  \hline
 \end{tabular}
 \caption{Statistics of the BQNC. NSW represents non-standard word.} \label{tab:stat}
\end{table}
We present the statistics of the BQNC in Table \ref{tab:stat}.
It comprises 929 sentences, 12.6K word tokens, and 767 non-standard word tokens.
As shown in Table \ref{tab:res_cate},
the corpus contains tokens of seven types of vocabulary and four types of variant form.
Whereas there exist fewer than 40 instances of neologisms/slang, dialect words, foreign words, and typographical errors,
each of the other category has more than 100 instances.
Our corpus contains a similar number of non-standard tokens to 
\newcite{kaji2014}'s Twitter corpus (1,831 sentences, 14.3K tokens, and 793 non-standard tokens)
and \newcite{osaki2017}'s Twitter corpus (1,405 sentences, 19.2K tokens, and 768 non-standard tokens).
The former follows the POS tags for the Japanese MA toolkit JUMAN
and the latter follows the authors own POS tags that extend NINJAL's SUW.

In the following subsections,
we evaluate the existing methods for MA and LN on the BQNC and
discuss correctly or incorrectly analyzed results.

\subsection{Systems}
We evaluated two existing methods.
First, we used MeCab 0.996 \cite{kudo2004},\footnote{\url{https://taku910.github.io/mecab/}}
which is a popular Japanese MA toolkit based on conditional random fields.
We used UniDicMA (unidic-cwj-2.3.0)\footnote{\url{https://unidic.ninjal.ac.jp/}}
as the analysis dictionary, which contains 
attribute information of 873K words 
and MeCab's parameters (word occurrence costs and transition costs) 
learned from annotated corpora, including the BCCWJ \cite{den2009}.

Second, we used our implementation of \newcite{sasano2013}'s joint MA and LN method.
They defined derivation rules 
to add new nodes in the word lattice of an input sentence built by 
their baseline system, JUMAN.
Specifically, they used the following rules:
(i) sequential voicing (\textit{rendaku}),
(ii) substitution with long sound symbols and lowercase kana,
(iii) insertion of long sound symbols and lowercase kana,
(iv) repetitive onomatopoeia (XYXY-form\footnote{``X'' and ``Y'' represent 
the same kana character(s) corresponding to one mora,
``Q'' represents a long consonant character ``\ja{っ}/\ja{ッ}'',  
``\textit{ri}'' represents a character ``\ja{り}/\ja{リ}'', 
and ``\textit{to}'' represents a character ``\ja{と}/\ja{ト}''.})
and (v) non-repetitive onomatopoeia (XQY\textit{ri}-form and XXQ\textit{to}-form).
For example, rule (iii) adds a node of \ja{冷たぁぁい} \textit{tsumet\=ai}
as a variant form of \ja{冷たい} \textit{tsumetai} `cold' and 
rule (iv) adds a node of \ja{うはうは} \textit{uhauha} `exhilarated' as an onomatopoeic adverb.
if the input sentences contain such character sequences.

The original implementation by \newcite{sasano2013} 
was an extension of JUMAN and followed JUMAN's POS tags.
To adapt their approach to the SUW, %criterion adopted in the BQNC,
we implemented their rules and used them to extend the first method of MeCab using UniDicMA.
We set the costs of the new nodes 
by copying the costs of their standard forms or the most frequent costs of the same-form onomatopoeia,
whereas \newcite{sasano2013} manually defined the costs of each type of new word.
We denote this method by MeCab+ER (Extension Rules).
Notably, we did not conduct any additional training to update the models' parameters for either methods.

\subsection{Overall Results}
Table \ref{tab:res_all} shows the overall performance,
that is, \textbf{P}recision, \textbf{R}ecall, and \textbf{F}${}_1$ score, of both methods 
for \textbf{SEG}mentation, \textbf{POS} tagging\footnote{We only evaluated top-level POS.} 
and \textbf{NOR}malization.\footnote{We regarded a predicted standard form as correct
if the prediction was equal to one of the gold standard forms.}
Compared with well-formed text domains,\footnote{For example, 
\newcite{kudo2004} achieved F${}_1$ of 98--99\% for segmentation and POS tagging in news domains.}
the relatively lower performance (F${}_1$ of 90--95\%) of both methods for segmentation and POS tagging
indicates the difficulty of accurate segmentation and tagging in UGT.
However, MeCab+ER outperformed MeCab by 2.5--2.9 F${}_1$ points because of the derivation rules.
Regarding the normalization performance of MeCab+ER,
the method achieved moderate precision but low recall,
which indicates its limited coverage for various variant forms in the dataset.

\begin{table}[t]
 \centering
 \footnotesize
 \begin{tabular}{l|rrr|rrr} \hline \hline
  Task & \multicolumn{3}{c|}{MeCab} & \multicolumn{3}{c}{MeCab+ER} \\
   & \multicolumn{1}{c}{P} & \multicolumn{1}{c}{R} & \multicolumn{1}{c|}{F} & \multicolumn{1}{c}{P} & \multicolumn{1}{c}{R} & \multicolumn{1}{c}{F} \\ \hline
  SEG & 89.2 & 95.1 & 92.1 & \textbf{93.5} & \textbf{96.5} & \textbf{95.0}\\
  POS & 87.5 & 93.3 & 90.3 & \textbf{91.4} & \textbf{94.3} & \textbf{92.8}\\
  NOR & --& --& --& 55.9 & 25.8 & 35.3 \\
  \hline
 \end{tabular}
 \caption{Overall performance} \label{tab:res_all}
\end{table}

\begin{table}[t]
 \centering
 \footnotesize
 \begin{tabular}{l|r|rr|rr} \hline \hline
  Category & \# & \multicolumn{2}{c|}{MeCab} & \multicolumn{2}{c}{MeCab+ER} \\
  & & SEG  & POS  & SEG  & POS \\ \hline
  Dialect words   &  23 & 91.3 & 78.3 & \textbf{95.7} & \textbf{82.6} \\
  Proper names    & 103 & 87.4 & 84.5 & \textbf{88.4} & \textbf{85.4} \\
  Onomatopoeia    & 218 & 79.8 & 73.4 & \textbf{87.2} & \textbf{77.1} \\
  Foreign words   &  14 & 78.6 & 78.6 & 78.6 & 78.6 \\
  Emoticons/AA    & 270 & 73.7 & 64.1 & 73.3 & 63.3 \\
  Interjections   & 174 & 64.9 & \textbf{53.5} & \textbf{72.4} & 48.9 \\
  Neologisms/Slang&  37 & 67.6 & 67.6 & 67.6 & 67.6 \\
  \hline
  Sound change var.$\!\!$& 419 & 50.6 & 47.5 & \textbf{82.6} & \textbf{76.4} \\
  Char type var.    & 248 & 71.0 & 62.9 & \textbf{78.2} & \textbf{69.4} \\
  Alternative rep. & 132 & 65.2 & 54.6 & \textbf{76.5} & \textbf{69.0} \\
  Typos            &  23 & 47.8 & 30.4 & 47.8 & 30.4 \\ \hline
  Non-gen/std total& 1565 & 68.9 & 61.9 & 79.6 & 70.4 \\ \hline \hline
  %Non-gen/std total& 1.6K & 68.9 & 61.9 & 79.6 & 70.4 \\ \hline \hline
  Standard forms of& \multirow{2}{*}{11K} & \multirow{2}{*}{98.9} & \multirow{2}{*}{97.7} & \multirow{2}{*}{98.9} & \multirow{2}{*}{97.7} \\
  general words & & & & & \\
  \hline
 \end{tabular}
 \caption{Recall for each category (SEG and POS)} \label{tab:res_cate}
\end{table}

\subsection{Results for Each Category}
Table \ref{tab:res_cate} shows 
the segmentation and POS tagging recall for both methods for each category.
In contrast to the sufficiently high performance for general words,
both methods performed worse for words of characteristic categories in UGT;
micro average recall was at most 79.6\% for segmentation and 70.4\% for POS tagging (``non-gen/std total'' column).
MeCab+ER outperformed MeCab particularly for
onomatopoeia, character type variants, alternative representations, and sound change variants.
The high scores for dialect words were probably because UniDicMA contains 
a large portion of (19 out of 23) dialect word tokens.
Interjection was a particularly difficult vocabulary type,
for which both methods recognized only approximately 50\% of the gold POS tags.
We guess that this is because the lexical variations of interjections are diverse; for example, 
there are many user-generated expressions that imitate various human voices, 
such as laughing, crying, and screaming.

Table \ref{tab:res_cate_n} shows the recall of MeCab+ER's normalization for each category.
The method correctly normalized tokens of alternative representations and
sound change variants with 30--40\% recall.
However, it completely failed to normalize character type variants 
not covered by the derivation rules and more irregular typographical errors.

\subsection{Analysis of the Segmentation Results} \label{sec:ana_seg}
\begin{table}[t]
 \centering
 \footnotesize
 \begin{tabular}{l|r|r} \hline \hline
  Category & \# & MeCab+ER \\ 
  \hline
  Sound change variants & 419 & 37.0 \\
  Character type variants & 248 & 0.0  \\
  Alternative representations& 132 & 32.6 \\
  Typographical errors  &  23 & 0.0  \\
  \hline
 \end{tabular}
 \caption{Recall for each category (normalization)} \label{tab:res_cate_n}
\end{table}

\begin{table}[t]
 \centering
 \footnotesize
 \begin{tabular}{l|rr} \hline \hline
  MeCab$\backslash$MeCab+ER & \multicolumn{1}{c}{T} & \multicolumn{1}{c}{F} \\ \hline
  T  & 11955 &  32 \\
  F &   200 & 413 \\
  \hline
 \end{tabular}
 \caption{Number of correct (T) or incorrect (F) segmentation for two methods} \label{tab:res_mat}
\end{table}
%TODO*2 for methods?

We performed error analysis of the segmentation results for the two methods.
Table \ref{tab:res_mat} shows a matrix of
the number of correct or incorrect segmentations for the methods for gold words.
There existed 32 tokens that only MeCab correctly segmented (T-F),
200 tokens that only MeCab+ER correctly segmented (F-T), and
413 tokens that both methods incorrectly segmented (F-F).

\begin{table*}[t]
 \centering
 \footnotesize
 \begin{tabular}{lllllll} \hline \hline
  & VT & Gold SEG\&SForms & Reading & Translation & MeCab result & MeCab+ER result \\
  \hline
  %T-F 
  (a) && \ja{はっ}$|$\ja{たり} & \textit{ha}Q$|$\textit{tari}  & paste and& \ja{はっ}$|$\ja{たり} & \red{\ja{はったり}} \\
  (b) && \ja{こら}$|$\ja{こら} & \textit{kora}$|$\textit{kora} & hey hey  & \ja{こら}$|$\ja{こら} & \red{\ja{こらこら}} \\
  \hline
  %F-T 
  (c) & S & \ja{しーかーも} \tiny{[\ja{しかも}]}   & \textit{sh\={\i}k\=amo}             & besides        & \red{\ja{しー}$|$\ja{かー}$|$\ja{も}} & \ja{しーかーも} \tiny{[\ja{しかも}]}  \\
  (d) & A & \ja{ぉぃら} \tiny{[\ja{おいら,オイラ}]} & \textit{oira}                    & I              & \red{\ja{ぉ}$|$\ja{ぃ}$|$\ja{ら}} & \ja{ぉぃら} \tiny{[\ja{おいら}]} \\
  (e) & S & \ja{んまぃ} \tiny{[\ja{美味い,旨い,うまい}]}   & \textit{mmai}                    & yummy          & \red{\ja{ん}$|$\ja{ま}$|$\ja{ぃ}}   & \ja{んまぃ} \tiny \red{[\ja{んまい}]}   \\
  (f) & C,A & \ja{も}$|$\ja{やきゅー} \tiny{[\ja{野球}]}  & \textit{mo}$|$\textit{yaky\=u}   & also, baseball & \red{\ja{もや}$|$\ja{きゅー}} & \ja{も}$|$\ja{やきゅー} \tiny \red{[\ja{やきゅう}]} \\
  (g) & S & \ja{たしーか} \tiny{[\ja{確か,たしか}]} \footnotesize$|$\ja{に}   & \textit{tash\={\i}ka}$|$\textit{ni} & surely  & \red{\ja{た}$|$\ja{し}$|$\ja{ー}$|$\ja{かに}} & \ja{たしーか} \tiny{[\ja{たしか}]} \footnotesize{$|$\ja{に}} \\
  (h) & & \ja{ふぅ〜〜ん}                     & \textit{f\=un}                   & hmm            & \red{\ja{ふぅ〜}$|$\ja{〜}$|$\ja{ん}} & \ja{ふぅ〜〜ん} \tiny \red{[\ja{ふん}]} \\
  \hline
  %F-F
  (i) & S & \ja{ませう〜} \tiny{[\ja{ましょう}]}     & \textit{mash\=o} & let's    & \red{\ja{ませ}$|$\ja{う}$|$\ja{〜}} & \red{\ja{ませ}$|$\ja{う〜}} \tiny \red{[\ja{う}]} \\
  (j) & C,S & \ja{けこーん} \tiny{[\ja{結婚}]}       & \textit{kek\=on} & marriage & \red{\ja{け}$|$\ja{こーん}} & \red{\ja{け}$|$\ja{こー}} \tiny \red{[\ja{こう}]} \footnotesize \red{$|$\ja{ん}} \\
  (k) & A & \ja{ください}$|\!$\ja{ｎｅ} \tiny{[\ja{ね}]}  & \textit{kudasai}$|$\textit{ne} & Won't you$\dots$? & \ja{ください}$|$\red{\ja{ｎ}$|$\ja{ｅ}} & \ja{ください}$|$\red{\ja{ｎ}$|$\ja{ｅ}} \\
  (l) && \ja{（＾へ＾）}           & & & \red{\ja{（}$|$\ja{＾}$|$\ja{へ}$|$\ja{＾}$|$\ja{）}} & \red{\ja{（}$|$\ja{＾}$|$\ja{へ}$|$\ja{＾}$|$\ja{）}} \\
  \multirow{2}{*}{(m)}$\!\!$ && \multirow{2}{*}{\ja{社割}} & \multirow{2}{*}{\textit{shawari}} & employee & \multirow{2}{*}{\red{\ja{社}$|$\ja{割}}} &\multirow{2}{*}{ \red{\ja{社}$|$\ja{割}}} \\
  &&&& discount \\
  (n) && \ja{ガルバディア}    & \textit{garubadhia} & Galbadia & \red{\ja{ガルバ}$|$\ja{ディア}} & \red{\ja{ガルバ}$|$\ja{ディア}} \\
  \hline
 \end{tabular}
 \caption{Segmentation and normalization results (shown in ``[]'') by MeCab and MeCab+ER.
 Incorrect results are written in gray.
 VT represents variant type.
 C, A, and S represent character type variant, alternative representation, and
 sound change variants, respectively.
 Gold SEG\&SForms represent the gold segmentation and gold standard forms (shown in ``[]'').
 } \label{tab:errors}
\end{table*}

In Table \ref{tab:errors}, we show the actual segmentation/normalization examples 
using the methods for the three cases;
the first, second, and third blocks show examples of T-F, F-T, and F-F cases, respectively.
First, out of 32 T-F cases, MeCab+ER incorrectly segmented tokens as onomatopoeia in 18 cases.
For example, (a) and (b) correspond to new nodes added 
by the rules for the XQY\textit{ri}-form and XYXY-form onomatopoeia,
respectively, even though (a) is a verb phrase 
%cosisting of a verb and an auxiliary verb 
and (b) is a repetition of interjections.

Second, out of 200 F-T cases that only MeCab+ER correctly segmented,
the method correctly normalized 119 cases, such as (c), (d), and 
%``たしーか'' (\textit{tash\={\i}ka}) 
the first word in (g), 
and incorrectly normalized 42 cases, such as (e) and 
%``やきゅー'' (\textit{yaky\=u}) 
the second word in (f).
The remaining 39 cases were tokens that required no normalization,
such as the first word 
%``も'' (\textit{mo}) 
in (f), 
%``に'' (\textit{ni}) 
the second word in (g), and (h).
The method correctly normalized simple examples of 
sound change variants (c: \ja{しーかーも} for \ja{しかも}) 
and alternative representations (d: \ja{ぉぃら} for \ja{おいら})
because of the substitution and insertion rules,
but failed to normalize character type variants (f: \ja{やきゅー} for \ja{野球}) 
and complicated sound change variants (e: \ja{んまぃ} for \ja{うまい}).

Third, out of 413 F-F cases,
148 tokens were complicated variant forms, including
a combination of historical kana orthography and the insertion of the long sound symbol (i),
%(i: ``ましょう'' $\rightarrow$ ``ませう'' $\rightarrow$ ``ませう〜''),
a combination of the character type variant and sound change variant (j),
%(j: ``結婚'' $\rightarrow$ ``けっこん'' $\rightarrow$ ``けこーん''), and
a variant written in \textit{romaji} 
%(k: ``ね'' $\rightarrow$ ``ｎｅ'').
(k).
The remaining 265 tokens were other unknown words, 
including emoticons (l), neologisms/slang (m),
and proper names (n).\footnote{\ja{社割} \textit{shawari} is an abbreviation of 
\ja{社員割引} \textit{shain waribiki} `employee discount'.
\ja{ガルバディア} `Galbadia' is an imaginary location 
name in the video game Final Fantasy.}

\begin{table}[t]
 \centering
 \footnotesize
 \begin{tabular}{l|r|l|r|l|r;{2pt/2pt}r} \hline \hline
  \multicolumn{2}{c|}{Total} & \multicolumn{4}{c|}{T-SEG} & F-SEG \\ \hline
  Gold & 767 & TP & 198 & FN & 58  & 511\\ \hline
  Pred & 354 & TP & 198 & FP & 99  &  57\\
  \hline
 \end{tabular}
 \caption{Detailed normalization results for MeCab+ER} \label{tab:res_norm}
\end{table}

\subsection{Analysis of the Normalization Results}
Table \ref{tab:res_norm} shows the detailed normalization results for MeCab+ER.
Among 767 non-standard words (Gold), the method correctly normalized 198 true positives (TP)
and missed 569 (58+511) false negatives (FN).
Similarly, among 354 predictions (Pred), the methods incorrectly normalized 156 (99+57) false positives (FP).
We further divided FN and FP according to whether they were correctly segmented (T-SEG) or not (F-SEG).

We do not show TP and FN examples here because we already introduced some examples in \S\ref{sec:ana_seg}.
Among the FP examples, some of them were not necessarily inappropriate results;
normalization between similar interjections and onomatopoeia was intuitively acceptable
(e.g., \ja{おお〜} was normalized to \ja{おお} \textit{\=o} `oh' and
\ja{サラサラ〜} was normalized to \ja{サラサラ} \textit{sarasara} `smoothly').
However, we assessed these as errors based on 
our criterion that interjections have no (non-)standard forms and 
the BCCWJ guidelines that regards
onomatopoeia with and without long sound insertion as different lemmas.

\subsection{Discussion}
The derivation rules used in MeCab+ER 
improved segmentation and POS tagging performance and 
contributed to the correct normalization of parts of variant forms,
but the overall normalization performance was limited to F${}_1$ of 35.3\%.

We classified the main segmentation and normalization errors into two types:
complicated variant forms and unknown words of specific vocabulary types
such as emoticons and neologisms/slang.
The effective use of linguistic resources may be required to build more accurate systems,
for example, discovering variant form candidates from large raw text similar to \cite{saito2017a},
and constructing/using term dictionaries of specific vocabulary types.

\section{Related Work}
\paragraph{UGT Corpus for MA and LN}~
\newcite{hashimoto2011} developed a Japanese blog corpus 
with morphological, grammatical, and sentiment information,
but it contains only 38 non-standard forms and 102 misspellings as UGT-specific examples.
% for evaluating LN performance.
\newcite{osaki2017} constructed a Japanese Twitter corpus annotated with
morphological information and standard word forms.
Although they published tweet URLs along with annotation information,
we could only restore parts of sentences because of the deletion of the original tweets.
%\footnote{We could only restore 33\% of the full data when we tried in April 2020.}
%\footnote{\url{https://github.com/tmu-nlp/TwitterCorpus}} 
\newcite{sasano2013,kaji2014,saito2014a,saito2017a} developed Japanese MA and LN methods 
for UGT, but most of their in-house data are not publicly available.
% \footnote{
% We contacted the first author of \cite{kaji2014} and he provided their corpus to us.
% We give a comparison with their corpus in \S\ref{sec:eval}.}
% Fairy Morphological Annotated Corpus \cite{hayashibe2017}.
%\footnote{\url{https://github.com/FairyDevicesRD/FairyMaCorpus}} 

For English LN, 
\newcite{han2011} constructed an English Twitter corpus 
%annotated corpus of 549 English tweets %(aka LexNorm 1.1)
and \newcite{yang2013} revised it as LexNorm 1.2.
\newcite{baldwin2015} constructed an English Twitter corpus (LexNorm2015)
%consisting of 3K training tweets and 2K test tweets, 
for the W-NUT 2015 text normalization shared task.
Both LexNorm 1.2 and LexNorm2015 have been used as benchmark datasets for LN systems
\cite{jin2015,vandergoot2019,dekker2020}.

For Chinese,
\newcite{li2008} published a dataset of formal-informal word pairs %501 
collected from Chinese webpages.
\newcite{wang2013} released a crowdsourced corpus constructed from %5.5K 
microblog posts on Sina Weibo.

\paragraph{Classification of Linguistic Phenomena in UGT}
To construct an MA dictionary, 
\newcite{nakamoto2000} classified unknown words  occurring in Japanese chat text
into contraction (e.g., \ja{すげー} for \ja{すごい} \textit{sugoi} `awesome'),
exceptional kana variant (e.g., \ja{こんぴゅーた} for \ja{コンピュータ} `computer'),
abbreviation, %(e.g., ``メアド'' for ``メールアドレス'' (mail address)),
typographical errors, filler, phonomime and phenomime, proper nouns, and other types.
\newcite{ikeda2010} classified
``peculiar expressions'' in Japanese blogs into 
visual substitution (e.g., \ja{わたＵ} for \ja{わたし} \textit{watashi} `me'),
sound change (e.g., \ja{でっかい} for \ja{でかい} \textit{dekai} `big'),
kana substitution (e.g., \ja{びたみん} for \ja{ビタミン} `vitamin'),
and other unknown words into similar categories to \newcite{nakamoto2000}.
% emoticons and ASCII art, emotional expressions (e.g., ``（笑）'' (slang indicating laughing)),
% phonomime and phenomime, and proper nouns.
\newcite{kaji2015} performed error analysis of Japanese MA methods on Twitter text.
They classified mis-segmented words into a dozen categories,
including spoken or dialect words, onomatopoeia,
interjections, emoticons/AA,
proper nouns, foreign words, misspelled words, and other non-standard word variants. 
\newcite{ikeda2010}'s classification of peculiar expressions is most similar to 
our types of variant forms and
\newcite{kaji2015}'s classification is most similar to our types of vocabulary 
(shown in Table \ref{sec:cate}),
whereas we provide more detailed definitions of categories and criteria for standard and non-standard forms.
Other work on Japanese MA and LN did not consider diverse phenomena in UGT \cite{sasano2013,saito2014a}.

For English, \newcite{han2011} classified ill-formed English words on Twitter into
extra/missing letters and/or number substitution (e.g., ``b4'' for ``before''),
slang (e.g., ``lol'' for ``laugh out loud'' ), and ``others''.
\newcite{vandergoot2018} defined a more comprehensive taxonomy with 14 categories
for a detailed evaluation of English LN systems.
It includes phrasal abbreviation (e.g., ``idk'' for ``I don't know''), 
repetition (e.g., ``soooo'' for ``so''), and phonetic transformation (e.g., ``hackd'' for ``hacked'').

For Chinese, 
\newcite{li2008} classified informal words in Chinese webpages into four types:
homophone (informal words with similar pronunciation to formal words, 
e.g.,  \zh{稀饭} $\langle$x\={\i}f\`an$\rangle$\footnote{Pinyin pronunciation is shown in ``$\langle\rangle$''.}
 ``rice gruel'' for \zh{喜欢} $\langle$x\v{\i}huan$\rangle$ ``like''),
abbreviation and acronym (e.g., GG for \zh{哥哥} $\langle$g\=ege$\rangle$ ``elder brother''), 
transliteration (informal words are transliteration of English translation of formal words,
e.g., 3Q $\langle$s\=anqiu$\rangle$ for \zh{谢谢} $\langle$xi\`exie$\rangle$ ``thank you''), 
and ``others''.
\newcite{wang2013} also classified informal words in Chinese microblog posts similar to \newcite{li2008}.

\paragraph{Methods for MA and LN}
In the last two decades, previous work has explored 
various rules and extraction methods for formal-informal word pairs 
to enhance Japanese MA and LN models for UGT.
% Japanese MA and LN methods for UGT have been developed in the last two decades.
\newcite{nakamoto2000} proposed an alignment method based on string similarity
between original and variant forms. 
%into a minimum conectivity-cost MA method for chat text.
\newcite{ikeda2010} automatically constructed normalization rules of peculiar expressions in blogs,
based on frequency, edit distance, and estimated accuracy improvements.
\newcite{sasano2013} defined derivation rules to recognize 
unknown onomatopoeia and variant forms of known words that frequently occur in webpages.
Their rules were also implemented in a recent MA toolkit Juman++ 
\cite{tolmachev2020} to handle unknown words.
% \newcite{kaji2014} developed a discriminative lattice traversal method for joint MA and LN
% using hand-crafted rules similar to \newcite{sasano2013}.
\newcite{saito2014a} estimated character-level alignment from manually annotated pairs of  
formal and informal words on Twitter. 
\newcite{saito2017a} extracted formal-informal word pairs from unlabeled Twitter data
based on semantic and phonetic similarity.
%Both work incorporated those normalization patterns in discriminative models for joint MA and LN.

For English and Chinese, various classification methods 
for normalization of informal words \cite{li2008,wang2013,han2011,jin2015,vandergoot2019}
have been developed based on, for example, string, phonetic, semantic similarity,
or co-occurrence frequency.
%spell checker's output, etc.
\newcite{qian2015} proposed a transition-based method with
append($x$), separate($x$), and separate\_and\_substitute($x$,$y$) operations
for the joint word segmentation, POS tagging, and normalization of Chinese microblog text.
\newcite{dekker2020} automatically generated pseudo training data from English raw tweets
using noise insertion operations %including hand-crafted rules
to achieve comparable performance without manually annotated data to an existing LN system.
%based on \newcite{vandergoot2018}'s taxonomy.
% They showed that an existing LN model \cite{vandergoot2019} trained on their pseudo training data
% achieved comparable performance to the same model trained on manually annotated data.

\section{Conclusion}
We presented a publicly available Japanese UGT corpus 
annotated with morphological and normalization information.
Our corpus enables the performance comparison of existing and future systems
and identifies the main remaining issues of MA and LN of UGT.
Experiments on our corpus demonstrated the
limited performance of the existing systems for non-general words and non-standard forms
%recall of at most 80\% for segmentation, 70\% for POS tagging, and F${}_1$ score of 35\% for normalization,
mainly caused by two types of difficult examples: 
complicated variant forms and unknown words of non-general vocabulary types.

In the future, we plan to
(1) expand the corpus by further annotating of 5--10 times more sentences 
for a more precise evaluation and 
(2) develop %a generation method for pseudo training data and 
a joint MA and LN method with high coverage.

\section*{Acknowledgments}
We would like to thank the anonymous reviewers and area chairs for their constructive comments.
We used the Balanced Corpus of Contemporary Written Japanese to construct our corpus.

\bibliography{anthology,mybib}
\bibliographystyle{acl_natbib}

% \appendix
% \section{Example Appendix}
% \label{sec:appendix}

\end{document}